\pdfoutput=1
\documentclass[final,12pt]{colt2023} 

\usepackage{times}

\usepackage{paralist,amsmath, amssymb}
\usepackage{algorithm}
\usepackage{algorithmic}
\usepackage{makecell}
\usepackage{multirow}

\def \K {\mathcal{K}}

\def \u {\mathbf{u}}
\def \x {\mathbf{x}}
\def \z {\mathbf{z}}
\def \c {\mathbf{c}}
\def \w {\mathbf{w}}
\def \ze {\mathbf{0}}

\def \v {\mathbf{v}}
\def \y {\mathbf{y}}

\DeclareMathOperator*{\argmin}{argmin}

\newtheorem{lem}{Lemma}

\newtheorem{assum}{Assumption}
\title{Improved Dynamic Regret for Online Frank-Wolfe}

\coltauthor{%
 \Name{Yuanyu Wan} \Email{wanyy@zju.edu.cn}\\
 \addr School of Software Technology, Zhejiang University
 \AND
 \Name{Lijun Zhang} \Email{zhanglj@lamda.nju.edu.cn}\\
 \addr National Key Laboratory for Novel Software Technology, Nanjing University
 \AND
 \Name{Mingli Song} \Email{brooksong@zju.edu.cn}\\
 \addr Shanghai Institute for Advanced Study, Zhejiang University
}

\begin{document}

\maketitle

\begin{abstract}%
To deal with non-stationary online problems with complex constraints, we investigate the dynamic regret of online Frank-Wolfe (OFW), which is an efficient projection-free algorithm for online convex optimization. It is well-known that in the setting of offline optimization, the smoothness of functions and the strong convexity of functions accompanying specific properties of constraint sets can be utilized to achieve fast convergence rates for the Frank-Wolfe (FW) algorithm. However, for OFW, previous studies only establish a dynamic regret bound of $O(\sqrt{T}(V_T+\sqrt{D_T}+1))$ by utilizing the convexity of problems, where $T$ is the number of rounds, $V_T$ is the function variation, and $D_T$ is the gradient variation. In this paper, we derive improved dynamic regret bounds for OFW by extending the fast convergence rates of FW from offline optimization to online optimization. The key technique for this extension is to set the step size of OFW with a line search rule. In this way, we first show that the dynamic regret bound of OFW can be improved to $O(\sqrt{T(V_T+1)})$ for smooth functions. Second, we achieve a better dynamic regret bound of $O(T^{1/3}(V_T+1)^{2/3})$ when functions are smooth and strongly convex, and the constraint set is strongly convex. Finally, for smooth and strongly convex functions with minimizers in the interior of the constraint set, we demonstrate that the dynamic regret of OFW reduces to $O(V_T+1)$, and can be further strengthened to $O(\min\{P_T^\ast,S_T^\ast,V_T\}+1)$ by performing a constant number of FW iterations per round, where $P_T^\ast$ and $S_T^\ast$ denote the path length and squared path length of minimizers, respectively.
\end{abstract}

 \section{Introduction}
Online convex optimization (OCO) is a powerful learning paradigm, which can be utilized to model a wide variety of machine learning problems, and is commonly formulated as a repeated game over a convex constraint set $\K\subseteq\mathbb{R}^d$ \citep{Hazan2016}. In each round $t=1,\dots,T$, a player first selects a feasible decision $\x_t\in\K$, and then suffers a loss $f_t(\x_t)$, where $f_t(\x):\K\mapsto\mathbb{R}$ is a convex function and could be selected in the adversarial way. The goal of the player is to minimize the cumulative loss $\sum_{t=1}^Tf_t(\x_t)$. A standard algorithm is online gradient descent (OGD) \citep{Zinkevich2003}, which usually has both theoretical and practical appeals. However, to ensure the feasibility of decisions, OGD needs to perform a projection operation per round, which could be computationally expensive in high-dimensional problems with complex constraints and thus limits its applications \citep{Hazan2012}. To tackle this issue, \citet{Hazan2012} propose the first projection-free algorithm for OCO, namely online Frank-Wolfe (OFW), by eschewing the projection operation with one iteration of the Frank-Wolfe (FW) algorithm \citep{FW-56,Revist_FW}, which is much more efficient for many complex constraint sets. 

Attracted by this computational advantage, there has been a growing research interest in developing and analyzing efficient projection-free online algorithms under different scenarios \citep{Garber16,kevy_smooth,Hazan20,Wan-AAAI-2021-C,Wan-ICML-2020,NeurIPS22-Wan,Wan-JMLR22,Kalhan2021,Garber-AISTATS21,Garber22,Zak_SC22,Zakaria21,ALT23-Zhou,OFW_Zhou}. However, most of them focus on minimizing the regret
\begin{equation}
\label{SR}
\sum_{t=1}^Tf_t(\x_t)-\min_{\x\in\K}\sum_{t=1}^Tf_t(\x)
\end{equation}
which compares the player against a fixed comparator, and thus cannot reflect the hardness of problems with non-stationary environments, where the best decision could be time-varying. To address this limitation, we investigate efficient projection-free algorithms with a more suitable metric called dynamic regret \citep{Zinkevich2003}
\begin{equation}
\label{DR1}
\sum_{t=1}^Tf_t(\x_t)-\sum_{t=1}^Tf_t(\x_t^\ast)
\end{equation} which compares the player against the minimizer $\x_t^\ast\in\argmin_{\x\in\K}f_t(\x)$ in each round $t$. Although a few studies \citep{Kalhan2021,OFW_Zhou} have proposed to minimize the dynamic regret with OFW \citep{Hazan2012}, current theoretical understandings of this way are limited. 
\begin{table}[t]
 \centering
 \caption{Summary of Convergence Rates of FW with $T$ iterations in Offline Optimization. Abbreviations: convex $\to$ cvx, strongly convex $\to$ scvx, smooth $\to$ sm.}
 \label{tab1}
 \begin{tabular}{|c|c|c|c|c|}
    \hline
    Reference & Function & $\K$ & \makecell{Location of Minimizers} & Convergence Rates\\
    \hline
    \citet{Revist_FW} & sm \& cvx & cvx & unrestricted & $O(1/T)$\\
    \hline
    \citet{Gaber_ICML_15} & sm \& scvx & scvx  & unrestricted & $O(1/T^2)$\\
    \hline
    \citet{Gaber_ICML_15} & sm \& scvx & cvx & the interior of $\K$ & $O(\exp(-T))$\\
    \hline
 \end{tabular}
\end{table}

To be precise, \citet{Kalhan2021} for the first time prove that OFW can attain a dynamic regret bound of $O(\sqrt{T}(V_T+\sqrt{D_T}+1))$ for smooth functions by directly applying the FW iteration to the function $f_t(\x)$, where $V_T=\sum_{t=2}^T\max_{\x\in\K}|f_{t}(\x)-f_{t-1}(\x)|$ and $D_T=\sum_{t=2}^T\|\nabla f_{t}(\x_t)-\nabla f_{t-1}(\x_{t-1})\|_2^2$ denote the function variation and the gradient variation, respectively. However, very recently, \citet{OFW_Zhou} show that OFW actually can attain the $O(\sqrt{T}(V_T+\sqrt{D_T}+1))$ dynamic regret bound by only utilizing the convexity of problems, which implies that the smoothness of functions is not appropriately exploited. Notice that in the setting of offline optimization, it is well-known that the smoothness of functions is essential for the convergence of the FW algorithm \citep{Revist_FW}. Moreover, if functions are smooth and strongly convex, the convergence rate of the FW algorithm can be further improved by utilizing the strong convexity of sets or the special location of the minimizers of functions (see Table \ref{tab1} for details) \citep{Gaber_ICML_15}. Thus, it is natural to ask whether these fast convergence rates of FW can be extended from offline optimization to online optimization for improving the dynamic regret of OFW.

In this paper, we provide an affirmative answer to the above question. Specifically, we first consider smooth functions and establish a dynamic regret bound of $O(\sqrt{T(V_T+1)})$ for OFW, which improves the existing $O(\sqrt{T}(V_T+\sqrt{D_T}+1))$ bound \citep{Kalhan2021,OFW_Zhou} by removing the dependence on $D_T$ and reducing the dependence on $V_T$. The insight behind these improvements is that by refining the existing analysis of OFW for smooth functions \citep{Kalhan2021}, the dynamic regret of OFW with a fixed step size $\sigma$ actually is $O(\sigma^{-1}(V_T+1)+\sigma T)$, which does not depend on $D_T$ and motivates us to minimize the dependence on $V_T$ by adjusting the step size. The key technical challenge is that the value of $V_T$ actually is unknowable in practice, which limits the choice of the step size. Inspired by previous studies on OCO \citep{Erven16,Zhang18_ader}, a standard way to address this limitation is to run multiple OFW, each with a different step size depending on an estimated $V_T$, and combine them via an expert-tracking algorithm. However, it requires $O(\log T)$ instances of OFW, which clearly increases the computational complexity and is unacceptable for real applications with a large $T$. By contrast, we adopt a simple line search rule \citep{Gaber_ICML_15} to select the step size of OFW, which can be implemented as efficient as the original OFW. 

Furthermore, if functions are smooth and strongly convex, we prove that the line search rule enables OFW to automatically reduce the dynamic regret to $O(T^{1/3}(V_T+1)^{2/3})$ over strongly convex sets, and $O(V_T+1)$ over convex sets in case the minimizers of functions lie in the interior of the set. These two improvements are analogous to the improvements in the convergence rate of FW for offline optimization. Finally, we demonstrate that under the same assumptions, the $O(V_T+1)$ dynamic regret can be further strengthened to $O(\min\{P_T^\ast,S_T^\ast,V_T\}+1)$ by performing a constant number of FW iterations per round, where $
P_T^\ast=\sum_{t=2}^T\|\x_t^\ast-\x^\ast_{t-1}\|_2$  and  $S_T^\ast=\sum_{t=2}^T\|\x_t^\ast-\x^\ast_{t-1}\|_2^2$ denote the path length and squared path length of minimizers, respectively. Notice that $P_T^\ast$, $S_T^\ast$, and $V_T$ reflect different aspects of the non-stationarity of environments, and thus are favored in different scenarios. Therefore, the strengthened bound achieves a \emph{best-of-three-worlds} guarantee. Moreover, this bound matches the best known dynamic regret bound achieved by a projection-based algorithm under the same assumptions \citep{Zhang17,L4DC:2021:Zhao}, which needs to perform a constant number of projected gradient descent iterations per round, and thus is less efficient than our algorithm for complex sets.

\section{Related Work}
\label{Sec_Rel}
In this section, we review related work on the  dynamic regret of projection-based and projection-free algorithms.

\subsection{Dynamic Regret of Projection-based Algorithms}
The pioneering work of \citet{Zinkevich2003} introduces a more general definition of the dynamic regret, which compares the player against any sequence of comparators
\begin{equation}
\label{DR2}
\sum_{t=1}^Tf_t(\x_t)-\sum_{t=1}^Tf_t(\u_t)
\end{equation}
where $\u_1,\dots,\u_T\in\K$, and shows that OGD with a constant step size can achieve a general dynamic regret bound of $O(\sqrt{T}(P_T+1))$ for convex functions, where $P_T=\sum_{t=2}^T\|\u_t-\u_{t-1}\|_2$. By running $O(\log T)$ instances of OGD with different step sizes and combining them with an expert-tracking algorithm, \citet{Zhang18_ader} improve the general dynamic regret to $O(\sqrt{T(P_T+1)})$, which has also been proved to be optimal for convex functions. The $\sqrt{T}$ part in this bound has been improved to some data-dependent terms for convex functions \citep{pmlr-v119-cutkosky20a} and smooth functions \citep{NIPS20-Zhao}. Furthermore, \citet{BabyCOLT} consider exponentially concave functions and strongly convex functions, and respectively establish a general dynamic regret bound of $\tilde{O}(d^{3.5}(T^{1/3}C_T^{2/3}+1))$ and an improved one of $\tilde{O}(d^2(T^{1/3}C_T^{2/3}+1))$,\footnote{The $\tilde{O}$ notation hides constant factors as well as polylogarithmic factors in $T$.} where $d$ is the dimensionality of decisions and $C_T=\sum_{t=2}^T\|\u_t-\u_{t-1}\|_1$.

Notice that the dynamic regret defined in (\ref{DR1}) can be regarded as the worst case of the general dynamic regret defined in (\ref{DR2}), which allows us to derive worst-case bounds by combining the above results with $\u_t=\x_t^\ast$. However, this way may not lead to the best results for the dynamic regret defined in (\ref{DR1}), since it ignores the specific property of $\x_t^\ast$. For this reason, there exist plenty of studies that focus on the dynamic regret defined in (\ref{DR1}). If each function $f_t(\x)$ is smooth and its minimizer $\x_t^\ast$ lies in the interior of $\K$, \citet{Yang16} show that OGD can achieve an $O(P_T^\ast+1)$ dynamic regret bound. When functions are smooth and strongly convex, \citet{Mokhtari16} establish the same dynamic regret bound for OGD. Moreover, when functions are smooth and strongly convex, and their minimizers lie in the interior of $\K$, \citet{Zhang17} propose the online multiple gradient descent (OMGD) algorithm, and improve the dynamic regret to $O(\min\{P_T^\ast,S_T^\ast\}+1)$. Besides, \citet{Besbes15} prove that a restarted variant of OGD can attain $O(T^{2/3}(V_T+1)^{1/3})$ and $\tilde{O}(\sqrt{T(V_T+1)})$ dynamic regret bounds for convex and strongly convex functions respectively, while they need to know the value of $V_T$ beforehand. By applying a strongly adaptive online learning framework \citep{FLH07,SAOL15}, in the case without the value of $V_T$, \citet{Zhang18} establish $\tilde{O}(T^{2/3}(V_T+1)^{1/3})$, $\tilde{O}(d\sqrt{T(V_T+1)})$, and $\tilde{O}(\sqrt{T(V_T+1)})$ dynamic regret bounds for convex functions, exponentially concave functions, and strongly convex functions, respectively. Under the same assumption as \citet{Zhang17}, \citet{L4DC:2021:Zhao} recently refine the dynamic regret of OMGD to $O(\min\{P_T^\ast,S_T^\ast,V_T\}+1)$.\footnote{The special location of the minimizer is only required to achieve the upper bound in terms of $S_T^\ast$.} They also demonstrate that a greedy strategy that sets $\x_{t+1}=\x_t^\ast$ can achieve the same dynamic regret bound without the strong convexity of functions.
\begin{table}[t]
 \centering
 \caption{Comparison of our results to previous studies. Here, we only summarize algorithms, which utilize at most $\tilde{O}(1)$ projections or FW iterations per round, and achieve dynamic regret bounds that are comparable with our results, i.e., depending on $V_T$. Abbreviations: convex $\to$ cvx, strongly convex $\to$ scvx, smooth $\to$ sm, the interior of $\K$ $\to$ $\text{inter}(\K)$, numbers of projections per round $\to$ \# PROJ, numbers of FW iterations per round $\to$ \# FW.} 
 \label{tab3}
 \begin{tabular}{|c|c|c|c|c|}
    \hline
      Assumptions& Reference &\# PROJ &\# FW& Dynamic Regret\\
    \hline
    \multirow{2}{*}{\makecell{$f_t(\cdot)$: cvx\\$\K$: cvx}} & \citet{Zhang18} & $\tilde{O}(1)$ & 0 & $\tilde{O}(T^{2/3}(V_T+1)^{1/3})$\\
    \cline{2-5}
    ~ & \citet{OFW_Zhou}& 0 & $1$ &$O(\sqrt{T}(V_T+\sqrt{D_T}+1))$\\ 
    \hline
    \multirow{2}{*}{\makecell{$f_t(\cdot)$: sm \& cvx \\$\K$: cvx}} & \citet{Kalhan2021}& 0 & 1 &$O(\sqrt{T}(V_T+\sqrt{D_T}+1))$\\ 
    \cline{2-5}
    ~& Theorem \ref{thm1}& 0 & 1 & $O(\sqrt{T(V_T+1)})$\\
    \hline
    \multirow{2}{*}{\makecell{$f_t(\cdot)$: sm \& scvx\\$\K$: scvx}} & \citet{L4DC:2021:Zhao}& $O(1)$ & 0 &$O(\min\{P_T^\ast,V_T\}+1)$\\ 
    \cline{2-5}
    ~& Theorem \ref{thm3} & 0 & $1$ & $O(T^{1/3}(V_T+1)^{2/3})$\\
    \hline
     \multirow{3}{*}{\makecell{$f_t(\cdot)$: sm \& scvx\\$\K$: cvx\\
     $\x_t^\ast\in\text{inter}(\K)$}} & \citet{L4DC:2021:Zhao}& $O(1)$ & 0 &$O(\min\{P_T^\ast,S_T^\ast,V_T\}+1)$\\ 
    \cline{2-5}
    ~& Theorem \ref{thm4}& 0 &$1$  & $O(V_T+1)$\\
    \cline{2-5}
    ~& Theorem \ref{thm1-OMFW}& 0& $O(1)$ & $O(\min\{P_T^\ast,S_T^\ast,V_T\}+1)$\\
    \hline
 \end{tabular}
\end{table}

\subsection{Dynamic Regret of Projection-free Algorithms}
However, the above algorithms require either the projection operation or more complicated computations, which cannot efficiently deal with complex constraints. To tackle this issue, \citet{Wan-AAAI-2021-B} propose an online algorithm that only utilizes FW iterations to update the decision, and establish ${O}(T^{2/3}(V_T+1)^{1/3})$ and $\tilde{O}(\sqrt{T(V_T+1)})$ dynamic regret bounds for convex and strongly convex functions, respectively. Unfortunately, the algorithm needs to utilize $\tilde{O}(T)$ and $\tilde{O}(T^2)$ FW iterations per round for achieving these two bounds, which actually suffers a similar computation complexity compared with the projection operation. By contrast, \citet{Kalhan2021} prove that by directly applying the FW iteration to the function $f_t(\x)$, OFW \citep{Hazan2012} can attain an $O(\sqrt{T}(V_T+\sqrt{D_T}+1))$ dynamic regret bound for smooth functions. Recently, \citet{OFW_Zhou} show that the same dynamic regret bound can be achieved by OFW without the smoothness of functions.

\subsection{Discussions}
As previously mentioned, if considering the regret defined in (\ref{SR}) rather than the dynamic regret, there exist plenty of projection-free online algorithms for different scenarios \citep{Garber16,kevy_smooth,Hazan20,Wan-AAAI-2021-C,Wan-ICML-2020,NeurIPS22-Wan,Wan-JMLR22,Garber-AISTATS21,Garber22,Zak_SC22,Zakaria21,ALT23-Zhou}. Although both the dynamic regret and projection-free algorithms have attracted much attention, OFW is the only efficient projection-free algorithm for minimizing the dynamic regret. In this paper, we improve the dynamic regret of OFW by extending the fast convergence rates of FW from offline optimization to online optimization. Moreover, it is worth noting that as summarized in Table \ref{tab3}, even compared with projection-based algorithms, our results can match the best $O(\min\{P_T^\ast,S_T^\ast,V_T\}+1)$ dynamic regret bound under the same assumptions, and exploit the smoothness of functions to improve the $\tilde{O}(T^{2/3}(V_T+1)^{1/3})$ dynamic regret bound achieved by only utilizing the convexity of functions.\footnote{In Appendix \ref{app_G}, we show that OGD \citep{Zinkevich2003} can also achieve the $O(\sqrt{T(V_T+1)})$ dynamic regret bound for smooth functions, which may be of independent interest.}

\section{Main Results}
We first revisit OFW for smooth functions, and then show the advantage of utilizing a line search rule. Finally, for smooth and strongly convex functions with minimizers in the interior of the set, we propose OFW with multiple updates to further strengthen the performance. 

\subsection{Revisiting Online Frank-Wolfe for Smooth Functions}
To minimize the dynamic regret, \citet{Kalhan2021} propose a specific instance of OFW, which first selects an arbitrary $\x_1\in\K$ and then iteratively updates the decision as follows
\begin{equation}
\label{eq_FWS}
\begin{split}
\v_t\in\argmin_{\x\in\K}\langle\nabla f_t(\x_t),\x\rangle,~\x_{t+1}=(1-\sigma)\x_t+\sigma \v_t
\end{split}
\end{equation}
where $\sigma\in[0,1]$ is a constant step size. To analyze the performance of this algorithm, \citet{Kalhan2021} introduce the following two assumptions.
\begin{assum}
\label{assum0}
All loss functions are $\alpha$-smooth, i.e., for any $t\in[T],\x\in\K,\y\in\K$, it holds that $f_t(\y)\leq f_t(\x)+\langle\nabla f_t(\x),\y-\x\rangle+\frac{\alpha}{2}\|\x-\y\|_2^2$.
\end{assum}
\begin{assum}
\label{assum2}
The diameter of the set $\K$ is bounded by $D$, i.e., for any $\x\in\K,\y\in\K$, it holds that $\|\x-\y\|_2\leq D$.
\end{assum}
Under these assumptions, \citet{Kalhan2021} establish the following lemma.
\begin{lem}
\label{worst-example}
(Lemma 1 of \citet{Kalhan2021})
Under Assumptions \ref{assum0} and \ref{assum2}, for $t=1,\cdots,T-1$, OFW ensures
\begin{equation}
\label{worst-example-eq}
\begin{split}
f_{t+1}(\x_{t+1})-f_{t+1}(\x_{t+1}^\ast)\leq&\max_{\x\in\K}|f_{t+1}(\x)-f_{t}(\x)|+(1-\sigma)(f_{t}(\x_{t})-f_{t}(\x_{t}^\ast))+\frac{3\alpha\sigma^2D^2}{2}\\
&+f_{t}(\x_{t}^\ast)-f_{t+1}(\x_{t+1}^\ast)+\sigma D\|\nabla f_{t+1}(\x_{t+1})-\nabla f_{t}(\x_{t})\|_2
\end{split}
\end{equation}
for the given step size $\sigma\in[0,1]$.
\end{lem}
By further utilizing the above lemma and setting $\sigma=1/\sqrt{T}$, \citet{Kalhan2021} achieve a dynamic regret bound of $O(\sqrt{T}(V_T+\sqrt{D_T}+1))$ for smooth functions. It is not hard to verify that the existence of $D_T$ in the bound is caused by the last term in the right side of (\ref{worst-example-eq}). 

In this paper, by analyzing OFW more carefully, we provide the following lemma for smooth functions, which removes the last term in the right side of (\ref{worst-example-eq}).
\begin{lem}
\label{good-lem1}
Under Assumptions \ref{assum0} and \ref{assum2}, for $t=1,\cdots,T-1$, OFW ensures
\begin{equation}
\label{good-example-eq}
\begin{split}
f_{t+1}(\x_{t+1})-f_{t+1}(\x_{t+1}^\ast)\leq&\max_{\x\in\K}|f_{t+1}(\x)-f_{t}(\x)|+(1-\sigma)(f_{t}(\x_{t})-f_{t}(\x_{t}^\ast))+\frac{\alpha\sigma^2D^2}{2}\\
&+f_{t}(\x_{t}^\ast)-f_{t+1}(\x_{t+1}^\ast)
\end{split}
\end{equation}
for the given step size $\sigma\in[0,1]$.
\end{lem}
Furthermore, based on Lemma \ref{good-lem1}, we can establish a simplified dynamic regret bound for OFW, which is presented in the following theorem.
\begin{theorem}
\label{thm0}
Under Assumptions \ref{assum0} and \ref{assum2}, OFW ensures
\begin{align*}
\sum_{t=1}^Tf_t(\x_t)-\sum_{t=1}^Tf_t(\x_t^\ast)\leq\frac{f_1(\x_1)-f_{T}(\x_{T}^\ast)+V_T}{\sigma}+\frac{\alpha\sigma(T-1)D^2}{2}
\end{align*}
where $V_T=\sum_{t=2}^T\max_{\x\in\K}|f_{t}(\x)-f_{t-1}(\x)|$.
\end{theorem}
\textbf{Remark.} From Theorem \ref{thm0}, OFW with $\sigma=1/\sqrt{T}$ actually enjoys a dynamic regret bound of $O(\sqrt{T}(V_T+1))$ for smooth functions, which is better than the $O(\sqrt{T}(V_T+\sqrt{D_T}+1))$ bound achieved by \citet{Kalhan2021} without any modification on the algorithm, and partially reflects the benefit of the smoothness of functions. More importantly, we notice that if $V_T$ is available beforehand, the dynamic regret bound of OFW can be further improved to $O(\sqrt{T(V_T+1)})$ by substituting $\sigma=O(\sqrt{V_T/T})$ into Theorem \ref{thm0}. However, $V_T$ is usually unknown in practical applications, which could also be the main reason why \citet{Kalhan2021} only utilize $\sigma=1/\sqrt{T}$. Inspired by previous studies \citep{Erven16,Zhang18_ader}, one may search the unknown value of $V_T$ by maintaining multiple instances of OFW, each with a different step size depending on an estimated $V_T$, and combining them with an expert-tracking algorithm. However, this way requires at least $O(\log T)$ instances of OFW, which clearly increases the computational complexity and is unacceptable for applications with a large $T$. In the following, we provide a more elegant way to achieve the $O(\sqrt{T(V_T+1)})$ bound.
\begin{algorithm}[t]
\caption{Online Frank-Wolfe with Line Search}
\label{alg1}
\begin{algorithmic}[1]
\STATE \textbf{Initialization:} $\x_1\in\K$
\FOR{$t=1,\cdots,T$}
\STATE Compute $\v_t\in\argmin_{\x\in\K}\langle\nabla f_t(\x_t),\x\rangle$
\STATE Compute $\sigma_t=\argmin_{\sigma\in[0,1]}\sigma\langle\nabla f_t(\x_t),\v_t-\x_t\rangle+\frac{\alpha\sigma^2}{2}\|\x_t-\v_t\|_2^2$
\STATE Update $\x_{t+1}=(1-\sigma_t)\x_t+\sigma_t \v_t$
\ENDFOR
\end{algorithmic}
\end{algorithm}
\subsection{Online Frank-Wolfe with Line Search}
Specifically, we utilize a line search rule to set the step size of OFW as follows 
\begin{equation}
\label{LS}
\sigma_t=\argmin_{\sigma\in[0,1]}\sigma\langle\nabla f_t(\x_t),\v_t-\x_t\rangle+\frac{\alpha\sigma^2}{2}\|\x_t-\v_t\|_2^2
\end{equation}
which is a common technique for setting the step size of FW in offline optimization \citep{Revist_FW,Gaber_ICML_15}, but usually is overlooked in OCO. By combining (\ref{eq_FWS}) with (\ref{LS}), we propose an algorithm called online Frank-Wolfe with line search, and the detailed procedures are summarized in Algorithm \ref{alg1}. 

We would like to emphasize that Algorithm \ref{alg1} can be implemented as efficient as the original OFW because the line search rule in (\ref{LS}) actually has a closed-form solution as 
\begin{equation*}
\sigma_t=\min\left\{\frac{\langle\nabla f_t(\x_t),\x_t-\v_t\rangle}{\alpha\|\x_t-\v_t\|_2^2},1\right\}.
\end{equation*}
To bound its dynamic regret, we first establish the following lemma for smooth functions, which generalizes Lemma \ref{good-lem1} from a fixed $\sigma$ to all possible $\sigma_\ast$.
\begin{lem}
\label{lem1}
Under Assumptions \ref{assum0} and \ref{assum2}, for $t=1,\cdots,T-1$, Algorithm \ref{alg1} ensures
\begin{equation}
\label{realeq-lem1}
\begin{split}
f_{t+1}(\x_{t+1})-f_{t+1}(\x_{t+1}^\ast)\leq& \max_{\x\in\K}|f_{t+1}(\x)-f_{t}(\x)|+(1-\sigma_\ast)(f_{t}(\x_{t})-f_{t}(\x_{t}^\ast))+\frac{\alpha\sigma_\ast^2D^2}{2}\\
&+f_{t}(\x_{t}^\ast)-f_{t+1}(\x_{t+1}^\ast)
\end{split}
\end{equation}
for all possible $\sigma_\ast\in[0,1]$.
\end{lem}
Based on the above lemma, we establish the following theorem.
\begin{theorem}
\label{thm1}
Under Assumptions \ref{assum0} and \ref{assum2}, Algorithm \ref{alg1} ensures
\begin{align*}
\sum_{t=1}^Tf_t(\x_t)-\sum_{t=1}^Tf_t(\x_t^\ast)\leq\sqrt{MT(V_T+M)}+\frac{\alpha D^2}{2}\sqrt{\frac{(V_T+M)T}{M}}
\end{align*}
where $V_T=\sum_{t=2}^T\max_{\x\in\K}|f_{t}(\x)-f_{t-1}(\x)|$ and $M=\max_{t\in[T],\x\in\K}\{2|f_t(\x)|\}$.
\end{theorem}
\textbf{Remark.} Theorem \ref{thm1} implies that by utilizing the line search rule, OFW can attain a dynamic regret bound of $O(\sqrt{T(V_T+1)})$ for smooth functions without the prior information of $V_T$. Compared with the $O(\sqrt{T}(V_T+\sqrt{D_T}+1))$ bound achieved by \citet{Kalhan2021}, this bound removes the dependence on $D_T$ and reduces the dependence on $V_T$.

Additionally, during the analysis of Lemma \ref{lem1} and Theorem \ref{thm1}, we realize that the convergence property of one FW iteration is critical for upper bounding $f_{t+1}(\x_{t+1})-f_{t+1}(\x_{t+1}^\ast)$, and then affects the dynamic regret. This motivates us to further investigate the dynamic regret of Algorithm \ref{alg1} in the following two special cases, in which the FW algorithm has been proved to converge faster \citep{Gaber_ICML_15}.
\subsubsection{Strongly Convex Functions and Sets}
To be precise, in the first case, we assume that these smooth functions are also strongly convex, and the decision set is strongly convex.
\begin{assum}
\label{sc-fun}
All loss functions are $\beta_f$-strongly convex, i.e., for any $t\in[T],\x\in\K,\y\in\K$, it holds that
\[f_t(\y)\geq f_t(\x)+\langle\nabla f_t(\x),\y-\x\rangle+\frac{\beta_f}{2}\|\x-\y\|_2^2.\]
\end{assum}
\begin{assum}
\label{assum-set}
The decision set is $\beta_K$-strongly convex, i.e., for any $\x\in\K,\y\in\K,\gamma\in[0,1]$ and $\z\in\mathbb{R}^d$ such that $\|\z\|_2=1$, it holds that
\[
\gamma\x+(1-\gamma)\y+\gamma(1-\gamma)\frac{\beta_K}{2}\|\x-\y\|_2^2\z \in\K.
\]
\end{assum}
By further incorporating these two assumptions, we establish the following lemma, which significantly improves Lemma \ref{lem1} by removing the $\frac{\alpha\sigma_\ast^2D^2}{2}$ term in the right side of (\ref{realeq-lem1}).
\begin{lem}
\label{lem1-thm3}
Under Assumptions \ref{assum0}, \ref{assum2}, and \ref{assum-set}, for $t=1,\cdots,T-1$, Algorithm \ref{alg1} ensures
\[
f_{t+1}(\x_{t+1})-f_{t+1}(\x_{t+1}^\ast)\leq \max_{\x\in\K}|f_{t+1}(\x)-f_{t}(\x)|+C_t(f_{t}(\x_{t})-f_{t}(\x_{t}^\ast))+f_{t}(\x_{t}^\ast)-f_{t+1}(\x_{t+1}^\ast)\]
where $C_t=\max\left\{1-\frac{\beta_K\|\nabla f_t(\x_t)\|_2}{8\alpha},\frac{1}{2}\right\}$.
\end{lem}
Based on Lemma \ref{lem1-thm3}, we establish the following lemma.
\begin{theorem}
\label{thm3}
Under Assumptions \ref{assum0}, \ref{assum2}, \ref{sc-fun}, and \ref{assum-set}, Algorithm \ref{alg1} ensures
\begin{align*}
\sum_{t=1}^Tf_t(\x_t)-\sum_{t=1}^Tf_t(\x_t^\ast)\leq \left(\frac{8\sqrt{2}\alpha}{\sqrt{\beta_f}\beta_K}(M+V_T)\right)^{2/3}T^{1/3}+2(M+V_T)
\end{align*}
where $V_T=\sum_{t=2}^T\max_{\x\in\K}|f_{t}(\x)-f_{t-1}(\x)|$ and $M=\max_{t\in[T],\x\in\K}\{2|f_t(\x)|\}$.
\end{theorem}
\textbf{Remark.} From Theorem \ref{thm3}, Algorithm \ref{alg1} can further exploit the strong convexity of functions and the set to achieve a dynamic regret bound of $O(T^{1/3}(V_T+1)^{2/3})$, which is better than the $O(\sqrt{T(V_T+1)})$ bound derived by only utilizing the smoothness of functions.
\subsubsection{Strongly Convex Functions with Special Minimizers}
Then, we proceed to consider the second case, in which all functions are strongly convex and their minimizers lie in the interior of the decision set. To this end, we further introduce the following assumption.
\begin{assum}
\label{interior}
There exists a minimizer $\x_t^\ast\in\argmin_{\x\in\K} f_t(\x)$ that lies in the interior of $\K$ for any $t\in[T]$, i.e., there exists a parameter $r>0$ such that if any $\x$ satisfies $\|\x-\x_t^\ast\|_2\leq r$, then $\x\in\K$.
\end{assum}
By replacing Assumption \ref{assum-set} in Lemma \ref{lem1-thm3} with Assumptions \ref{sc-fun} and \ref{interior}, we establish the following lemma.
\begin{lem}
\label{lem1-thm4}
Under Assumptions \ref{assum0}, \ref{assum2}, \ref{sc-fun}, and \ref{interior}, for $t=1,\cdots,T-1$, Algorithm \ref{alg1} ensures
\[
f_{t+1}(\x_{t+1})-f_{t+1}(\x_{t+1}^\ast)\leq \max_{\x\in\K}|f_{t+1}(\x)-f_{t}(\x)|+C(f_{t}(\x_{t})-f_{t}(\x_{t}^\ast))+f_{t}(\x_{t}^\ast)-f_{t+1}(\x_{t+1}^\ast)
\]
where $C=1-\frac{\beta_f\tilde{r}^2}{4\alpha D^2}$, $\tilde{r}=\min\left\{r,\frac{\sqrt{2}\alpha D^2}{\sqrt{\beta_f M}}\right\}$, and $M=\max_{t\in[T],\x\in\K}\{2|f_t(\x)|\}$.
\end{lem}
Lemma \ref{lem1-thm4} improves Lemma \ref{lem1-thm3} by replacing the variable $C_t$ with a constant $C$. Moreover, we derive the following theorem.
\begin{theorem}
\label{thm4}
Under Assumptions \ref{assum0}, \ref{assum2}, \ref{sc-fun}, and \ref{interior}, Algorithm \ref{alg1} ensures
\begin{align*}
\sum_{t=1}^Tf_t(\x_t)-\sum_{t=1}^Tf_t(\x_t^\ast)\leq \frac{4\alpha(M+V_T)D^2}{\beta_f\tilde{r}^2}
\end{align*}
where $M=\max_{t\in[T],\x\in\K}\{2|f_t(\x)|\}$, $\tilde{r}=\min\left\{r,\frac{\sqrt{2}\alpha D^2}{\sqrt{\beta_f M}}\right\}$, and $V_T=\sum_{t=2}^T\max_{\x\in\K}|f_{t}(\x)-f_{t-1}(\x)|$.
\end{theorem}
\textbf{Remark.} Theorem \ref{thm4} implies that Algorithm \ref{alg1} can achieve an $O(V_T+1)$ dynamic regret bound over convex sets by further exploiting the strong convexity of functions and the special location of minimizers. This bound is better than the $O(T^{1/3}(V_T+1)^{2/3})$ bound given by Theorem \ref{thm3}, which implies that the special location of minimizers is more useful for OFW than the strong convexity of sets.

\subsection{Online Frank-Wolfe with Multiple Updates}
Furthermore, we note that under the same assumptions as in our Theorem \ref{thm4}, previous studies \citep{L4DC:2021:Zhao} have achieved an $O(\min\{P_T^\ast,S_T^\ast,V_T\}+1)$ dynamic regret bound, which is tighter than the $O(V_T+1)$ bound achieved by Algorithm \ref{alg1}. To fill this gap, we strengthen Algorithm \ref{alg1} by performing a small number of FW iterations per round. Intuitively, more FW iterations will result in smaller $f_{t+1}(\x_{t+1})-f_{t+1}(\x_{t+1}^\ast)$, and thus reduce the dynamic regret. Specifically, the detailed procedures are outlined in Algorithm \ref{alg2}, which is named as online Frank-Wolfe with multiple updates and enjoys the following theoretical guarantee.
\begin{algorithm}[t]
\caption{Online Frank-Wolfe with Multiple Updates}
\label{alg2}
\begin{algorithmic}[1]
\STATE \textbf{Input:} the iteration number $K$ per round
\STATE \textbf{Initialization:} $\x_1\in\K$
\FOR{$t=1,\cdots,T$}
\STATE $\z_{t+1}^0=\x_t$
\FOR{$i=0,\cdots,K-1$}
\STATE Compute $\v_t^i\in\argmin_{\x\in\K}\langle\nabla f_t(\z_{t+1}^i),\x\rangle$
\STATE Compute $\sigma_t^i=\argmin_{\sigma\in[0,1]}\sigma\langle\nabla f_t(\z_{t+1}^i),\v_t^i-\z_{t+1}^i\rangle+\frac{\alpha\sigma^2}{2}\|\z_{t+1}^i-\v_t^i\|_2^2$
\STATE Update $\z_{t+1}^{i+1}=(1-\sigma_t^i)\z_{t+1}^{i}+\sigma_t^i \v_t^i$
\ENDFOR
\STATE $\x_{t+1}=\z_{t+1}^{K}$
\ENDFOR
\end{algorithmic}
\end{algorithm}
\begin{theorem}
\label{thm1-OMFW}
Under Assumptions \ref{assum0}, \ref{assum2}, \ref{sc-fun}, and \ref{interior}, by setting $K=\left\lceil\frac{\ln(\beta_f/4\alpha)}{\ln C}\right\rceil$ where $C=1-\frac{\beta_f\tilde{r}^2}{4\alpha D^2}$, $\tilde{r}=\min\left\{r,\frac{\sqrt{2}\alpha D^2}{\sqrt{\beta_f M}}\right\}$, and $M=\max_{t\in[T],\x\in\K}\{2|f_t(\x)|\}$, Algorithm \ref{alg2} ensures 
\begin{align*}
\sum_{t=1}^Tf_t(\x_t)-\sum_{t=1}^Tf_t(\x_t^\ast)\leq &\min\left\{\frac{4\alpha(M+V_T)}{4\alpha-\beta_f},2GD+2GP_T^\ast,\alpha D^2+2\alpha S_T^\ast\right\}
\end{align*}
where ${G}=\max_{t\in[T],\x\in\K}\|\nabla f_t(\x_t)\|_2$, $V_T=\sum_{t=2}^T\max_{\x\in\K}|f_{t}(\x)-f_{t-1}(\x)|$, $
P_T^\ast=\sum_{t=2}^T\|\x_t^\ast-\x^\ast_{t-1}\|_2$, and $S_T^\ast=\sum_{t=2}^T\|\x_t^\ast-\x^\ast_{t-1}\|_2^2$.
\end{theorem}
\textbf{Remark.} From Theorem \ref{thm1-OMFW}, for smooth and strongly convex functions with minimizers in the interior of the set, our Algorithm \ref{alg2} can also achieve the $O(\min\{P_T^\ast,S_T^\ast,V_T\}+1)$ dynamic regret bound by using a constant number of FW iterations per round, which strengthens the $O(V_T+1)$ bound achieved by only using one FW iteration per round.

\section{Theoretical Analysis}
In this section, we only prove Lemma \ref{lem1}, Theorem \ref{thm1}, and Theorem \ref{thm4}. The omitted proofs are provided in the appendix. Compared with the existing analysis of OFW \citep{Kalhan2021}, the main novelty of our analysis is based on an appropriate way to incorporate the convergence property of one FW iteration \citep{Gaber_ICML_15}.
\subsection{Proof of Lemma \ref{lem1}}
We first decompose $f_{t+1}(\x_{t+1})-f_{t+1}(\x_{t+1}^\ast)$ as follows
\begin{equation}
\label{lem1-eq1}
\begin{split}
f_{t+1}(\x_{t+1})-f_{t+1}(\x_{t+1}^\ast)=&f_{t+1}(\x_{t+1})-f_{t}(\x_{t+1})+f_{t}(\x_{t+1})-f_{t}(\x_{t}^\ast)+f_{t}(\x_{t}^\ast)-f_{t+1}(\x_{t+1}^\ast)\\
\leq&\max_{\x\in\K}|f_{t+1}(\x)-f_t(\x)|+f_{t}(\x_{t+1})-f_{t}(\x_{t}^\ast)+f_{t}(\x_{t}^\ast)-f_{t+1}(\x_{t+1}^\ast).
\end{split}
\end{equation}
Then, we only need to analyze $f_t(\x_{t+1})-f_{t}(\x_t^\ast)$. Moreover, because $\x_{t+1}$ is generated by applying one FW iteration on $f_t(\x_{t})$, it is natural to bound $f_t(\x_{t+1})-f_{t}(\x_t^\ast)$ by exploiting the convergence property of one FW iteration.

Specifically, by using the smoothness of $f_t(\x)$, we have
\begin{align*}
f_t(\x_{t+1})-f_{t}(\x_t^\ast)\leq&f_t(\x_{t})-f_{t}(\x_t^\ast)+\langle\nabla f_t(\x_t),\x_{t+1}-\x_{t}\rangle+\frac{\alpha}{2}\|\x_{t+1}-\x_{t}\|_2^2.
\end{align*}
Moreover, according to Algorithm \ref{alg1}, we have 
\[\x_{t+1}-\x_{t}=\sigma_t(\v_t-\x_t).\]
Combining the above two equations, we have
\begin{equation}
\label{lem1-eq2}
\begin{split}
f_t(\x_{t+1})-f_{t}(\x_t^\ast)\leq&f_t(\x_{t})-f_{t}(\x_t^\ast)+\sigma_t\langle\nabla f_t(\x_t),\v_t-\x_t\rangle+\frac{\alpha\sigma^2_t}{2}\|\v_t-\x_t\|_2^2\\
\overset{(\ref{LS})}{\leq}&f_t(\x_{t})-f_{t}(\x_t^\ast)+\sigma_\ast\langle\nabla f_t(\x_t),\v_t-\x_t\rangle+\frac{\alpha\sigma_\ast^2}{2}\|\v_t-\x_t\|_2^2
\end{split}
\end{equation}
for all possible $\sigma_\ast\in[0,1]$.

Because of $\v_t\in\argmin_{\x\in\K}\langle\nabla f_t(\x_t),\x\rangle$ in Algorithm \ref{alg1}, we also have
\begin{equation}
\label{eq_convex_loss}
\begin{split}
\langle\nabla f_t(\x_t),\v_t-\x_t\rangle\leq&\langle\nabla f_t(\x_t),\x_t^\ast-\x_t\rangle\\
\leq&f_t(\x_t^\ast)-f_t(\x_t)
\end{split}
\end{equation}
where the last inequality is due to the convexity of $f_t(\x)$.

Then, combining (\ref{lem1-eq2}) with (\ref{eq_convex_loss}), we have
\begin{equation}
\label{lem1-eq2-post}
\begin{split}
f_t(\x_{t+1})-f_{t}(\x_t^\ast)\leq&(1-\sigma_\ast)(f_t(\x_{t})-f_{t}(\x_t^\ast))+\frac{\alpha\sigma_\ast^2}{2}\|\v_t-\x_t\|_2^2\\
\leq&(1-\sigma_\ast)(f_t(\x_{t})-f_{t}(\x_t^\ast))+\frac{\alpha\sigma_\ast^2D^2}{2}
\end{split}
\end{equation}
where the last inequality is due to Assumption \ref{assum2}. 

Finally, we complete this proof by substituting  (\ref{lem1-eq2-post}) into (\ref{lem1-eq1}).

\subsection{Proof of Theorem \ref{thm1}}
Let $R_T^D=\sum_{t=1}^Tf_t(\x_t)-\sum_{t=1}^Tf_t(\x_t^\ast)$. We first have
\begin{equation}
\label{base_case}
\begin{split}
R_T^D=&\sum_{t=2}^Tf_t(\x_t)-\sum_{t=2}^Tf_t(\x_t^\ast)+f_1(\x_1)-f_1(\x_1^\ast)
\end{split}
\end{equation}
By using Lemma \ref{lem1}, for any $\sigma_\ast\in[0,1]$, we have
\begin{align*}
R_T^D\leq&f_1(\x_1)-f_1(\x_1^\ast)+\sum_{t=2}^T\max_{\x\in\K}|f_{t}(\x)-f_{t-1}(\x)|+\sum_{t=2}^T(1-\sigma_\ast)(f_{t-1}(\x_{t-1})-f_{t-1}(\x_{t-1}^\ast))\\
&+\sum_{t=2}^T\left(f_{t-1}(\x_{t-1}^\ast)-f_{t}(\x_{t}^\ast)+\frac{\alpha\sigma_\ast^2D^2}{2}\right)\\
=&f_1(\x_1)-f_1(\x_1^\ast)+V_T-(1-\sigma_\ast)(f_{T}(\x_{T})-f_{T}(\x_{T}^\ast))+(1-\sigma_\ast)R_T^D\\
&+f_{1}(\x_{1}^\ast)-f_{T}(\x_{T}^\ast)+\frac{\alpha(T-1)\sigma_\ast^2D^2}{2}\\
\leq&f_1(\x_1)-f_{T}(\x_{T}^\ast)+V_T+\frac{\alpha(T-1)\sigma_\ast^2D^2}{2}+(1-\sigma_\ast)R_T^D
\end{align*}
where the last inequality is due to $f_{T}(\x_{T})-f_{T}(\x_{T}^\ast)\geq0$ and $\sigma_\ast\in[0,1]$. 

Then, we have
\begin{align*}
\sigma_\ast R_T^D\leq& f_1(\x_1)-f_{T}(\x_{T}^\ast)+V_T+\frac{\alpha(T-1)\sigma_\ast^2D^2}{2}\\
\leq&M+V_T+\frac{\alpha(T-1)\sigma_\ast^2D^2}{2}
\end{align*}
where the last inequality is due to the definition of $M$.

By dividing both sides of the above inequality with $\sigma_\ast$ and setting $\sigma_\ast=\sqrt{\frac{V_T+M}{MT}}$, we have
\begin{align*}
R_T^D\leq&\frac{M+V_T}{\sigma_\ast}+\frac{\alpha(T-1)\sigma_\ast D^2}{2}\\
\leq&\sqrt{MT(V_T+M)}+\frac{\alpha D^2}{2}\sqrt{\frac{(V_T+M)T}{M}}.
\end{align*}

\subsection{Proof of Theorem \ref{thm4}}
Let $R_T^D=\sum_{t=1}^Tf_t(\x_t)-\sum_{t=1}^Tf_t(\x_t^\ast)$. Combining (\ref{base_case}) with Lemma \ref{lem1-thm4}, we have
\begin{align*}
R_T^D\leq&f_1(\x_1)-f_1(\x_1^\ast)+\sum_{t=2}^T\max_{\x\in\K}|f_{t}(\x)-f_{t-1}(\x)|+\sum_{t=2}^TC(f_{t-1}(\x_{t-1})-f_{t-1}(\x_{t-1}^\ast))\\
&+\sum_{t=2}^T\left(f_{t-1}(\x_{t-1}^\ast)-f_{t}(\x_{t}^\ast)\right)\\
=&f_1(\x_1)-f_{1}(\x_{1}^\ast)+V_T-C(f_{T}(\x_{T})-f_{T}(\x_{T}^\ast))+CR_T^D+f_1(\x_1^\ast)-f_{T}(\x_{T}^\ast)\\
\leq&f_1(\x_1)-f_{T}(\x_{T}^\ast)+V_T+CR_T^D
\end{align*}
where the last inequality is due to $f_{T}(\x_{T})-f_{T}(\x_{T}^\ast)\geq0$ and $C\geq0$.

From the above inequality, it is easy to verify that
\begin{align*}
R_T^D\leq\frac{f_1(\x_1)-f_{T}(\x_{T}^\ast)+V_T}{1-C}=\frac{4\alpha D^2(f_1(\x_1)-f_{T}(\x_{T}^\ast)+V_T)}{\beta_f\tilde{r}^2}\leq\frac{4\alpha D^2(M+V_T)}{\beta_f\tilde{r}^2}
\end{align*}
where the equality is due to the definition of $C$ and the last inequality is due to the definition of $M$.

\section{Conclusion and Future Work}
In this paper, we first improve the dynamic regret bound of the OFW algorithm for smooth functions from $O(\sqrt{T}(V_T+\sqrt{D_T}+1))$ to $O(\sqrt{T(V_T+1)})$. Second, if functions are smooth and strongly convex, we establish a tighter bound of $O(T^{1/3}(V_T+1)^{2/3})$ for OFW over strongly convex sets. Finally, for smooth and strongly convex functions with minimizers in the interior of the constraint set, we show that the dynamic regret of OFW reduces to $O(V_T+1)$, and can be further strengthened to $O(\min\{P_T^\ast,S_T^\ast,V_T\}+1)$ by utilizing a constant number of FW iterations per round.

Note that all our results of OFW require the smoothness of functions. A natural open problem is whether our results can be extended into the non-smooth case. From our current analysis, the convergence of FW is critical for the dynamic regret of OFW. However, for non-smooth functions, FW is not guaranteed to converge in general. Thus, it seems highly non-trivial to control the dynamic regret of OFW or its variants in the non-smooth case. Moreover, the line search rule in (\ref{LS}) is only valid for smooth functions. Although one may replace it with $\sigma_t=\argmin_{\sigma\in[0,1]}f_t((1-\sigma)\x_t+\sigma\v_t)$ in the non-smooth case, the time complexity is unclear and could be much higher than that of (\ref{LS}). For these reasons, we leave this problem as a future work.

Another potential limitation of this paper is that the dynamic regret defined in (\ref{DR1}) could be too pessimistic as it compares the player against the minimizer of each round. To address this limitation, very recently, we have proposed an algorithm with an $O(T^{3/4}(P_T+1)^{1/4})$ upper bound on the general dynamic regret defined in (\ref{DR2}) for convex functions, by utilizing $O(\log T)$ FW iterations per round \citep{Wang-Arxiv23}. However, it is still unclear whether the upper bound can be further improved without or with additional assumptions on functions and sets, which will be investigated in the future.

\acks{This work was partially supported by the National Natural Science Foundation of China (61976186, U20B2066), the Key Research and Development Program of Zhejiang Province (No. 2023C03192), the Starry Night Science Fund of Zhejiang University Shanghai Institute for Advanced Study (No. SN-ZJU-SIAS-001), and the Fundamental Research Funds for the Central Universities (2021FZZX 001-23, 226-2023-00048). The authors would also like to thank the anonymous reviewers for their helpful comments.}

\bibliography{ref}

\newpage
\appendix

\section{Proof of Theorem \ref{thm0}}
Let $R_T^D=\sum_{t=1}^Tf_t(\x_t)-\sum_{t=1}^Tf_t(\x_t^\ast)$. We first notice that (\ref{base_case}) in the proof of Theorem \ref{thm1} still holds for OFW. Combining it with Lemma \ref{good-lem1}, we have
\begin{align*}
R_T^D\leq&f_1(\x_1)-f_1(\x_1^\ast)+\sum_{t=2}^T\max_{\x\in\K}|f_{t}(\x)-f_{t-1}(\x)|+\sum_{t=2}^T(1-\sigma)(f_{t-1}(\x_{t-1})-f_{t-1}(\x_{t-1}^\ast))\\
&+\sum_{t=2}^T\left(f_{t-1}(\x_{t-1}^\ast)-f_{t}(\x_{t}^\ast)+\frac{\alpha\sigma^2D^2}{2}\right)\\
=&f_1(\x_1)-f_{T}(\x_{T}^\ast)+V_T-(1-\sigma)(f_{T}(\x_{T})-f_{T}(\x_{T}^\ast))+(1-\sigma)R_T^D+\frac{\alpha(T-1)\sigma^2D^2}{2}\\
\leq&f_1(\x_1)-f_{T}(\x_{T}^\ast)+V_T+\frac{\alpha(T-1)\sigma^2D^2}{2}+(1-\sigma)R_T^D
\end{align*}
where the last inequality is due to $f_{T}(\x_{T})-f_{T}(\x_{T}^\ast)\geq0$ and $\sigma\in[0,1]$. 

From the above inequality, it is easy to verify that
\begin{align*}
\sigma R_T^D\leq& f_1(\x_1)-f_{T}(\x_{T}^\ast)+V_T+\frac{\alpha(T-1)\sigma^2D^2}{2}.
\end{align*}
Finally, we complete this proof by dividing both sides of the above inequality with $\sigma$.

\section{Proof of Theorem \ref{thm3}}
Combining (\ref{base_case}) with Lemma \ref{lem1-thm3}, we have
\begin{align*}
&\sum_{t=1}^Tf_t(\x_t)-\sum_{t=1}^Tf_t(\x_t^\ast)\\
\leq&f_1(\x_1)-f_1(\x_1^\ast)+\sum_{t=2}^T\max_{\x\in\K}|f_{t}(\x)-f_{t-1}(\x)|+\sum_{t=2}^TC_{t-1}(f_{t-1}(\x_{t-1})-f_{t-1}(\x_{t-1}^\ast))\\
&+\sum_{t=2}^T\left(f_{t-1}(\x_{t-1}^\ast)-f_{t}(\x_{t}^\ast)\right)\\
=&f_1(\x_1)-f_{T}(\x_{T}^\ast)+V_T+\sum_{t=1}^{T-1}C_{t}(f_{t}(\x_{t})-f_{t}(\x_{t}^\ast))\\
\leq&f_1(\x_1)-f_{T}(\x_{T}^\ast)+V_T+\sum_{t=1}^{T}C_{t}(f_{t}(\x_{t})-f_{t}(\x_{t}^\ast))
\end{align*}
where the second inequality is due to $f_{T}(\x_{T})-f_{T}(\x_{T}^\ast)\geq0$ and $C_T\geq0$.

Then, for brevity, we define 
\begin{equation}
\label{bad_set}
\mathcal{S}_T=\left\{t\in[T]\left|\frac{1}{2}\leq1-\frac{\beta_K\|\nabla f_t(\x_t)\|_2}{8\alpha}\right.\right\}.
\end{equation}
From the above two equations, we have
\begin{equation}
\label{pre_final0}
\begin{split}
&\sum_{t\in\mathcal{S}_T}\frac{\beta_K\|\nabla f_t(\x_t)\|_2}{8\alpha}(f_{t}(\x_{t})-f_{t}(\x_{t}^\ast))+\sum_{t\in[T]\setminus\mathcal{S}_T}\frac{1}{2}(f_{t}(\x_{t})-f_{t}(\x_{t}^\ast))\\
\leq& f_1(\x_1)-f_{T}(\x_{T}^\ast)+V_T\leq M+V_T
\end{split}
\end{equation}
where the last inequality is due to the definition of $M$.

To analyze the first term in the left side of (\ref{pre_final0}), we further note that as proved by \citet{Gaber_ICML_15}, if a function $f(\x):\K\mapsto\mathbb{R}$ is $\beta_f$-strongly convex, we have
\begin{equation}
\label{eq-scv}
\|\nabla f(\x)\|_2\geq \sqrt{\frac{\beta_f}{2}}\sqrt{f(\x)-f(\x_\ast)}
\end{equation}
for any $\x\in\K$ and $\x_\ast=\argmin_{\x\in\K}f(\x)$. 

Because of Assumption \ref{sc-fun}, we have
\begin{equation}
\label{pre_final1}
\begin{split}
\sum_{t\in\mathcal{S}_T}(f_{t}(\x_{t})-f_{t}(\x_{t}^\ast))^{3/2}\overset{(\ref{eq-scv})}{\leq}&\sum_{t\in\mathcal{S}_T}\sqrt{\frac{2}{{\beta_f}}}\|\nabla f_t(\x_t)\|_2(f_{t}(\x_{t})-f_{t}(\x_{t}^\ast))\\
\overset{(\ref{pre_final0})}{\leq}&\frac{8\sqrt{2}\alpha}{\sqrt{\beta_f}\beta_K}(M+V_T).
\end{split}
\end{equation}
Finally, it is not hard to verify that
\begin{align*}
\sum_{t=1}^Tf_t(\x_t)-\sum_{t=1}^Tf_t(\x_t^\ast)\overset{(\ref{bad_set})}{=}&\sum_{t\in\mathcal{S}_T}(f_{t}(\x_{t})-f_{t}(\x_{t}^\ast))+\sum_{t\in[T]\setminus\mathcal{S}_T}(f_{t}(\x_{t})-f_{t}(\x_{t}^\ast))\\
\overset{(\ref{pre_final0})}{\leq}&\sum_{t\in\mathcal{S}_T}(f_{t}(\x_{t})-f_{t}(\x_{t}^\ast))+2(M+V_T)\\
\leq&\left(\sum_{t\in\mathcal{S}_T}(f_{t}(\x_{t})-f_{t}(\x_{t}^\ast))^{3/2}\right)^{2/3}T^{1/3}+2(M+V_T)\\
\overset{(\ref{pre_final1})}{\leq}&\left(\frac{8\sqrt{2}\alpha}{\sqrt{\beta_f}\beta_K}(M+V_T)\right)^{2/3}T^{1/3}+2(M+V_T)
\end{align*}
where the second inequality is due to Hölder's inequality and $\left|\mathcal{S}_T\right|\leq T$.

\section{Proof of Theorem \ref{thm1-OMFW}}
We introduce the following lemma.
\begin{lem}
\label{lem-app1}
Under Assumptions \ref{assum0}, \ref{assum2}, \ref{sc-fun}, and \ref{interior}, for any $t\in[T]$, Algorithm \ref{alg2} ensures
\begin{align*}
f_t(\x_{t+1})-f_t(\x_t^\ast)\leq C^K(f_t(\x_{t})-f_{t}(\x_t^\ast))
\end{align*}
where $C=1-\frac{\beta_f\tilde{r}^2}{4\alpha D^2}$ and $\tilde{r}=\min\left\{r,\frac{\sqrt{2}\alpha D^2}{\sqrt{\beta_f M}}\right\}$.
\end{lem}
Combining (\ref{lem1-eq1}) with Lemma \ref{lem-app1}, for $t=1,\cdots,T-1$, we have
\begin{align*}
f_{t+1}(\x_{t+1})-f_{t+1}(\x_{t+1}^\ast)\leq&\max_{\x\in\K}|f_{t+1}(\x)-f_t(\x)|+C^K(f_{t}(\x_{t})-f_{t}(\x_{t}^\ast))\\
&+f_{t}(\x_{t}^\ast)-f_{t+1}(\x_{t+1}^\ast).
\end{align*}
Then, similar to the proof of Theorem \ref{thm4}, by substituting the above inequality into (\ref{base_case}), it is not hard to verify that
\begin{align*}
\sum_{t=1}^Tf_t(\x_t)-\sum_{t=1}^Tf_t(\x_t^\ast)
\leq&f_1(\x_1)-f_{T}(\x_{T}^\ast)+V_T+C^K\left(\sum_{t=1}^Tf_t(\x_t)-\sum_{t=1}^Tf_t(\x_t^\ast)\right)
\end{align*}
which implies that
\begin{equation}
\label{V_T}
\sum_{t=1}^Tf_t(\x_t)-\sum_{t=1}^Tf_t(\x_t^\ast)\leq \frac{f_1(\x_1)-f_{T}(\x_{T}^\ast)+V_T}{1-C^K}\leq\frac{4\alpha(M+V_T)}{4\alpha-\beta_f}
\end{equation}
where the last inequality is due to the definition of $M$ and 
\begin{equation}
\label{Small_C}
C^K\leq C^{\frac{\ln(\beta_f/4\alpha)}{\ln C}}=\frac{\beta_f}{4\alpha}.
\end{equation}
To further derive dynamic regret bounds depending on $P_T^\ast$ and $S_T^\ast$, we introduce nice properties of strongly convex functions and smooth functions. First, as proved by \citet{Gaber_ICML_15}, if a function $f(\x)$ is $\beta_f$-strongly convex over $\K$, we have
\begin{equation}
\label{eq2-scv}
f(\x)-f(\x_\ast)\geq\frac{\beta_f}{2}\|\x-\x_\ast\|_2^2
\end{equation}
for any $\x\in\K$ and $\x_\ast=\argmin_{\x\in\K}f(\x)$. 

Second, from Assumption \ref{assum0}, $f_t(\x)$ is $\alpha$-smooth over $\K$, which satisfies that
\begin{equation}
\label{eq-new-smooth}
\begin{split}
f_t(\x)-f_t(\x_t^\ast)\leq&\langle\nabla f_t(\x_t^\ast),\x-\x_t^\ast\rangle+\frac{\alpha}{2}\|\x_t^\ast-\x\|_2^2\\
=&\frac{\alpha}{2}\|\x_t^\ast-\x\|_2^2
\end{split}
\end{equation}
for any $\x\in\K$, where the last equality is due to the fact that Assumption \ref{interior} also implies $\nabla f_t(\x_t^\ast)=\ze$.

By using these two properties, for any $t\in[T]$, we have
\begin{equation}
\label{eq-converge}
\begin{split}
\|\x_{t+1}-\x_t^\ast\|_2^2\overset{(\ref{eq2-scv})}{\leq}&\frac{2}{\beta_f}\left(f_t(\x_{t+1})-f_t(\x_t^\ast)\right)\leq\frac{2C^K}{\beta_f}\left(f_t(\x_{t})-f_t(\x_t^\ast)\right)\\
\overset{(\ref{eq-new-smooth})}{\leq}&\frac{\alpha C^K}{\beta_f}\|\x_{t}-\x_t^\ast\|_2^2\overset{(\ref{Small_C})}{\leq}\frac{1}{4}\|\x_t-\x_t^\ast\|_2^2
\end{split}
\end{equation}
where the second inequality is due to Lemma \ref{lem-app1}.

Then, because of Assumption \ref{assum2}, it is not hard to verify that
\begin{equation}
\label{sum-eq-converge1}
\begin{split}
\sum_{t=1}^T\|\x_{t}-\x_t^\ast\|_2\leq&\|\x_{1}-\x_1^\ast\|_2+\sum_{t=2}^T\|\x_{t}-\x_t^\ast\|_2\\
\leq&D+\sum_{t=2}^T\|\x_{t}-\x_{t-1}^\ast\|_2+\sum_{t=2}^T\|\x_{t}^\ast-\x_{t-1}^\ast\|_2\\
\overset{(\ref{eq-converge})}{\leq}&D+\frac{1}{2}\sum_{t=2}^T\|\x_{t-1}-\x_{t-1}^\ast\|_2+P_T^\ast
\end{split}
\end{equation}
which implies that 
\begin{equation}
\label{pre_PT}
\begin{split}
\sum_{t=1}^T\|\x_{t}-\x_t^\ast\|_2
\leq&2D+2P_T^\ast.
\end{split}
\end{equation}
Now, we achieve the dynamic regret bound depending on $P_T^\ast$ as follows
\begin{equation}
\label{eq-PT}
\begin{split}
\sum_{t=1}^Tf_t(\x_{t})-\sum_{t=1}^Tf_t(\x_{t}^\ast)\leq&\sum_{t=1}^T\langle\nabla f_t(\x_{t}),\x_{t}-\x_{t}^\ast\rangle\leq\sum_{t=1}^TG\|\x_{t}^\ast-\x_{t}^\ast\|_2\\
\overset{(\ref{pre_PT})}{\leq}&2GD+2GP_T^\ast
\end{split}
\end{equation}
where the second inequality is due to the definition of $G$.

Furthermore, similar to (\ref{sum-eq-converge1}), we also have
\begin{equation*}
\begin{split}
\sum_{t=1}^T\|\x_{t}-\x_t^\ast\|_2^2\leq&\|\x_{1}-\x_1^\ast\|_2^2+\sum_{t=2}^T\|\x_{t}-\x_t^\ast\|_2^2\\
\leq&D^2+2\sum_{t=2}^T\|\x_{t}-\x_{t-1}^\ast\|_2^2+2\sum_{t=2}^T\|\x_{t}^\ast-\x_{t-1}^\ast\|_2^2\\
\overset{(\ref{eq-converge})}{\leq}&D^2+\frac{1}{2}\sum_{t=2}^T\|\x_{t-1}-\x_{t-1}^\ast\|_2^2+2S_T^\ast
\end{split}
\end{equation*}
which implies that 
\begin{equation}
\label{pre_ST}
\begin{split}
\sum_{t=1}^T\|\x_{t}-\x_t^\ast\|_2^2
\leq&2D^2+4S_T^\ast.
\end{split}
\end{equation}
Then, we have
\begin{equation}
\label{eq-ST}
\begin{split}
\sum_{t=1}^Tf_t(\x_{t})-\sum_{t=1}^Tf_t(\x_{t}^\ast)\overset{(\ref{eq-new-smooth})}{\leq}\frac{\alpha}{2}\sum_{t=1}^T\|\x_{t}-\x_{t}^\ast\|_2^2\overset{(\ref{pre_ST})}{\leq}\alpha D^2+2\alpha S_T^\ast.
\end{split}
\end{equation}
Finally, by combining (\ref{V_T}), (\ref{eq-PT}), and (\ref{eq-ST}), we complete this proof.

\section{Proof of Lemma \ref{good-lem1}}
We first note that (\ref{lem1-eq1}) and (\ref{eq_convex_loss}) in the proof of Lemma \ref{lem1} still hold for OFW. So, we only need to analyze $f_t(\x_{t+1})-f_{t}(\x_t^\ast)$. According to (\ref{eq_FWS}), we have \[
\x_{t+1}-\x_{t}=\sigma(\v_t-\x_t).\]
Then, combining this equality with the smoothness of $f_t(\x)$, we have
\begin{equation}
\label{lem1-eq2-worst}
\begin{split}
f_t(\x_{t+1})-f_{t}(\x_t^\ast)\leq&f_t(\x_{t})-f_{t}(\x_t^\ast)+\sigma\langle\nabla f_t(\x_t),\v_t-\x_t\rangle+\frac{\alpha\sigma^2}{2}\|\v_t-\x_t\|_2^2\\
\leq&f_t(\x_{t})-f_{t}(\x_t^\ast)+\sigma\langle\nabla f_t(\x_t),\v_t-\x_t\rangle+\frac{\alpha\sigma^2D^2}{2}\\
\overset{(\ref{eq_convex_loss})}{\leq}&(1-\sigma)(f_t(\x_{t})-f_{t}(\x_t^\ast))+\frac{\alpha\sigma^2D^2}{2}
\end{split}
\end{equation}
where the second inequality is due to Assumption \ref{assum2}.

Finally, we complete this proof by substituting (\ref{lem1-eq2-worst}) into (\ref{lem1-eq1}).

\section{Proof of Lemma \ref{lem1-thm3}}
Following the proof of Lemma \ref{lem1}, we still utilize (\ref{lem1-eq1}) and (\ref{lem1-eq2}). The main difference is to establish a tighter bound for $f_t(\x_{t+1})-f_t(\x_{t}^\ast)$ by further using Assumption \ref{assum-set}. Specifically, we first define 
\begin{equation*}
\c_t=\frac{1}{2}(\x_t+\v_t), \w_t\in\argmin_{\|\w\|_2\leq1}\langle\nabla f_t(\x_t),\w\rangle, \text{ and }
\v_t^\prime=\c_t+\frac{\beta_K}{8}\|\x_t-\v_t\|_2^2\w_t.
\end{equation*}
According to Assumption \ref{assum-set}, it is easy to verify that \[\v_t^\prime\in\K.\]
Moreover, because of $\v_t\in\argmin_{\x\in\K}\langle\nabla f_t(\x_t),\x\rangle$ in Algorithm \ref{alg1}, we have
\begin{equation}
\label{eq_convex_loss2}
\begin{split}
\langle\nabla f_t(\x_t),\v_t-\x_t\rangle\leq&\langle\nabla f_t(\x_t),\v_t^\prime-\x_t\rangle\\
=&\frac{1}{2}\langle\nabla f_t(\x_t),\v_t-\x_t\rangle+\frac{\beta_K\|\x_t-\v_t\|_2^2}{8}\langle\nabla f_t(\x_t),\w_t\rangle\\
\overset{(\ref{eq_convex_loss})}{\leq}&-\frac{1}{2}(f_t(\x_t)-f_t(\x_t^\ast))+\frac{\beta_K\|\x_t-\v_t\|_2^2}{8}\langle\nabla f_t(\x_t),\w_t\rangle\\
=&-\frac{1}{2}(f_t(\x_t)-f_t(\x_t^\ast))-\frac{\beta_K\|\x_t-\v_t\|_2^2}{8}\|\nabla f_t(\x_t)\|_2
\end{split}
\end{equation}
where the last equality is due to
\begin{equation}
\label{supp-eq1}
\begin{split}
\langle\nabla f_t(\x_t),\w_t\rangle=\min_{\|\w\|_2\leq1}\langle\nabla f_t(\x_t),\w\rangle=-\|\nabla f_t(\x_t)\|_2.
\end{split}
\end{equation}
Substituting (\ref{eq_convex_loss2}) into (\ref{lem1-eq2}), for all possible $\sigma_\ast\in[0,1]$, we have
\begin{equation}
\label{lem1-thm2-eq1}
\begin{split}
&f_t(\x_{t+1})-f_{t}(\x_t^\ast)\\
\leq&\left(1-\frac{\sigma_\ast}{2}\right)(f_t(\x_{t})-f_{t}(\x_t^\ast))+\frac{\alpha\sigma^2_\ast}{2}\|\v_t-\x_t\|_2^2-\frac{\sigma_\ast\beta_K\|\x_t-\v_t\|_2^2}{8}\|\nabla f_t(\x_t)\|_2\\
=&\left(1-\frac{\sigma_\ast}{2}\right)(f_t(\x_{t})-f_{t}(\x_t^\ast))+\left(\alpha\sigma^2_\ast-\frac{\sigma_\ast\beta_K\|\nabla f_t(\x_t)\|_2}{4}\right)\frac{\|\x_t-\v_t\|_2^2}{2}.
\end{split}
\end{equation}
Then, if $\alpha\leq\frac{\beta_K\|\nabla f_t(\x_t)\|_2}{4}$, by combining $\sigma_\ast=1$ with (\ref{lem1-thm2-eq1}), we have
\begin{equation}
\label{lem1-thm2-eq2}
\begin{split}
f_t(\x_{t+1})-f_{t}(\x_t^\ast)
\leq\frac{1}{2}(f_t(\x_{t})-f_{t}(\x_t^\ast)).
\end{split}
\end{equation}
Otherwise, by combining $\sigma_\ast=\frac{\beta_K\|\nabla f_t(\x_t)\|_2}{4\alpha}$  with (\ref{lem1-thm2-eq1}), we have
\begin{equation}
\label{lem1-thm2-eq3}
\begin{split}
f_t(\x_{t+1})-f_{t}(\x_t^\ast)\leq&
\left(1-\frac{\beta_K\|\nabla f_t(\x_t)\|_2}{8\alpha}\right)(f_t(\x_{t})-f_{t}(\x_t^\ast)).
\end{split}
\end{equation}
Combining (\ref{lem1-thm2-eq2}) and (\ref{lem1-thm2-eq3}) with the definition of $C_t$, we have
\begin{equation}
\label{lem1-thm2-eq4}
\begin{split}
f_t(\x_{t+1})-f_{t}(\x_t^\ast)
\leq C_t(f_t(\x_{t})-f_{t}(\x_t^\ast)).
\end{split}
\end{equation}
Finally, we complete this proof by substituting (\ref{lem1-thm2-eq4}) into (\ref{lem1-eq1}).

\section{Proof of Lemma \ref{lem1-thm4}}
Note that Algorithm \ref{alg2} with $K=1$ reduces to Algorithm \ref{alg1}. Therefore, Lemma \ref{lem-app1} with $K=1$ also holds for Algorithm \ref{alg1}, which implies that 
\begin{equation*}
\begin{split}
f_t(\x_{t+1})-f_{t}(\x_t^\ast)
\leq C(f_t(\x_{t})-f_{t}(\x_t^\ast)).
\end{split}
\end{equation*}
By substituting the above inequality into (\ref{lem1-eq1}), we complete this proof.

\section{Proof of Lemma \ref{lem-app1}}
This lemma can be derived from the convergence rate of FW under Assumptions \ref{assum0}, \ref{assum2}, \ref{sc-fun}, and \ref{interior} (see Section 4.2 of \citet{Gaber_ICML_15} for the details). For the sake of completeness, we include the detailed proof here.

We first define
\begin{equation*}
\w_t^i\in\argmin_{\|\w\|_2\leq1}\langle\nabla f_t(\z_{t+1}^i),\w\rangle \text{ and } \tilde{\v}_t^i=\x_{t}^\ast+\tilde{r}\w_t^i
\end{equation*} 
where $\tilde{\v}_t^i$ also belongs to $\K$ since $\x_t^\ast$ lies in the interior of $\K$ and $\tilde{r}\leq r$.
 
Then, because of $\v_t^i\in\argmin_{\x\in\K}\langle\nabla f_t(\z_{t+1}^i),\x\rangle$ in Algorithm \ref{alg2}, we have
\begin{equation}
\label{eq1-thm4-app}
\begin{split}
\langle\nabla f_t(\z_{t+1}^i),\v_t^i-\z_{t+1}^i\rangle\leq&\langle\nabla f_t(\z_{t+1}^i),\tilde{\v}_t^i-\z_{t+1}^i\rangle\\
=&\langle\nabla f_t(\z_{t+1}^i),\x_t^\ast-\z_{t+1}^i\rangle+\tilde{r}\langle\nabla f_t(\z_{t+1}^i),\w_t^i\rangle\\
\leq&-\tilde{r}\|\nabla f_t(\z_{t+1}^i)\|_2
\end{split}
\end{equation}
where the last inequality is due to \[\langle\nabla f_t(\z_{t+1}^i),\w_t^i\rangle=\min_{\|\w\|_2\leq1}\langle\nabla f_t(\z_{t+1}^i),\w\rangle
=-\|\nabla f_t(\z_{t+1}^i)\|_2\]
and \[\langle\nabla f_t(\z_{t+1}^i),\x_t^\ast-\z_{t+1}^i\rangle\leq f_t(\x_t^\ast)-f_t(\z_{t+1}^i)\leq0.\]
Combining the smoothness of $f_t(\x)$ with $\z_{t+1}^{i+1}-\z_{t+1}^{i}=\sigma_t^i(\v_t^i-\z_{t+1}^{i})$ in Algorithm \ref{alg2}, for all possible $\sigma_\ast\in[0,1]$, we have
\begin{equation}
\label{appedn-111-pre}
\begin{split}
f_t(\z_{t+1}^{i+1})-f_{t}(\x_t^\ast)\leq&f_t(\z_{t+1}^{i})-f_{t}(\x_t^\ast)+\sigma_t^i\langle\nabla f_t(\z_{t+1}^{i}),\v_{t}^{i}-\z_{t+1}^{i}\rangle+\frac{\alpha(\sigma_t^i)^2}{2}\|\v_{t}^{i}-\z_{t+1}^{i}\|_2^2\\
\leq&f_t(\z_{t+1}^{i})-f_{t}(\x_t^\ast)+\sigma_\ast\langle\nabla f_t(\z_{t+1}^{i}),\v_{t}^{i}-\z_{t+1}^{i}\rangle+\frac{\alpha\sigma_\ast^2}{2}\|\v_{t}^{i}-\z_{t+1}^{i}\|_2^2\\
\leq&f_t(\z_{t+1}^{i})-f_{t}(\x_t^\ast)+\sigma_\ast\langle\nabla f_t(\z_{t+1}^{i}),\v_{t}^{i}-\z_{t+1}^{i}\rangle+\frac{\alpha\sigma_\ast^2D^2}{2}\\
\overset{(\ref{eq1-thm4-app})}{\leq}&f_t(\z_{t+1}^{i})-f_{t}(\x_t^\ast)-\sigma_\ast\tilde{r}\|\nabla f_t(\z_{t+1}^i)\|_2+\frac{\alpha\sigma_\ast^2D^2}{2}\\
\overset{(\ref{eq-scv})}{\leq}&f_t(\z_{t+1}^{i})-f_{t}(\x_t^\ast)-\sigma_\ast\tilde{r}\sqrt{\frac{\beta_f(f_t(\z_{t+1}^i)-f_t(\x_t^\ast))}{2}}+\frac{\alpha\sigma_\ast^2D^2}{2}
\end{split}
\end{equation}
where the second inequality is due to the line search rule in Algorithm \ref{alg2} and the third inequality is due to Assumption \ref{assum2}.

Moreover, we set $\sigma_\ast$ as
\[
0\leq\sigma_\ast=\frac{\tilde{r}\sqrt{\beta_f(f_t(\z_{t+1}^i)-f_t(\x_t^\ast))}}{\sqrt{2}\alpha D^2}\leq\frac{\tilde{r}\sqrt{\beta_fM}}{\sqrt{2}\alpha D^2}{\leq}1
\]
where the last inequality is due to the definition of $\tilde{r}$.

Combining this $\sigma_\ast$ with (\ref{appedn-111-pre}), we have
\begin{equation*}
\begin{split}
f_t(\z_{t+1}^{i+1})-f_{t}(\x_t^\ast)\leq&f_t(\z_{t+1}^{i})-f_{t}(\x_t^\ast)-\frac{\tilde{r}^2\beta_f(f_t(\z_{t+1}^{i})-f_{t}(\x_t^\ast))}{4\alpha D^2}=C(f_t(\z_{t+1}^{i})-f_{t}(\x_t^\ast))
\end{split}
\end{equation*}
where the last equality is due to the definition of $C$.

Finally, from the above inequality, it is easy to verify that
\begin{align*}
f_t(\x_{t+1})-f_{t}(\x_t^\ast)=&f_t(\z_{t+1}^{K})-f_{t}(\x_t^\ast)\leq C^K(f_t(\z_{t+1}^0)-f_{t}(\x_t^\ast))\\
=&C^K(f_t(\x_{t})-f_{t}(\x_t^\ast)).
\end{align*}

\section{Dynamic Regret of OGD for Smooth Functions}
\label{app_G}
As discussed in Section \ref{Sec_Rel}, our $O(\sqrt{T(V_T+1)})$ dynamic regret for smooth functions is not achieved by previous studies even using projection-based algorithms. Here, we notice that OGD \citep{Zinkevich2003} actually also enjoys this bound. Specifically, the detailed procedures of OGD are summarized in Algorithm \ref{ogd}, where the step size is simply set as $1/\alpha$ by following the common choice in offline optimization \citep{Bubeck_survey}. Then, we establish an $O(\sqrt{T(V_T+1)})$ dynamic regret bound for Algorithm \ref{ogd} in the following theorem.

\begin{theorem}
\label{thm_ogd}
Under Assumptions \ref{assum0} and \ref{assum2}, Algorithm \ref{ogd} ensures
\begin{align*}
\sum_{t=1}^Tf_t(\x_t)-\sum_{t=1}^Tf_t(\x_t^\ast)\leq M+V_T+\sqrt{2\alpha D^2(T-1)(M+V_T)}
\end{align*}
where $V_T=\sum_{t=2}^T\max_{\x\in\K}|f_{t}(\x)-f_{t-1}(\x)|$ and $M=\max_{t\in[T],\x\in\K}\{2|f_t(\x)|\}$.
\end{theorem}
\begin{algorithm}[t]
\caption{Online Gradient Descent}
\label{ogd}
\begin{algorithmic}[1]
\STATE \textbf{Initialization:} $\x_1\in\K$
\FOR{$t=1,\cdots,T$}
\STATE Compute $\x_{t+1}^\prime=\x_t-\frac{1}{\alpha} \nabla f_t(\x_t)$
\STATE Compute $\x_{t+1}=\argmin_{\x\in\K}\|\x-\x_{t+1}^\prime\|_2^2$
\ENDFOR
\end{algorithmic}
\end{algorithm}
\begin{proof}
We start this proof by introducing the following lemma.
\begin{lem}
\label{lem1_OGD}
(Lemma 3.6 of \citet{Bubeck_survey})
Let $f(\cdot):\K\mapsto\mathbb{R}$ be a convex and $\alpha$-smooth function, and $\x^\prime=\argmin_{\x\in\K}\left\|\x-\frac{1}{\alpha}\nabla f(\x)\right\|_2^2$. Then, for any $\x\in\K,\y\in\K$, we have
\begin{equation*}
f(\x^\prime)-f(\y)\leq\alpha\langle\x-\x^\prime,\x-\y\rangle-\frac{\alpha}{2}\|\x-\x^\prime\|_2^2.
\end{equation*}
\end{lem}
From Lemma \ref{lem1_OGD}, it is easy to verify that
\begin{equation}
\label{OGD_eq2}
f_t(\x_{t+1})-f_t(\x_t)\leq-\frac{\alpha}{2}\|\x_{t+1}-\x_{t}\|_2^2
\end{equation}
and
\begin{equation}
\label{OGD_eq3}
\begin{split}
f_t(\x_{t+1})-f_t(\x_t^\ast)
\leq&\alpha\|\x_t-\x_{t+1}\|_2\|\x_t-\x_t^\ast\|_2.
\end{split}
\end{equation}
Then, combining with Assumption \ref{assum2}, we have
\begin{equation}
\label{OGD_eq4}
\begin{split}
f_t(\x_{t+1})-f_t(\x_t^\ast)\overset{(\ref{OGD_eq2})}{\leq}&f_t(\x_{t})-f_t(\x_t^\ast)-\frac{\alpha}{2}\|\x_{t+1}-\x_{t}\|_2^2\\
\overset{(\ref{OGD_eq3})}{\leq}&f_t(\x_{t})-f_t(\x_t^\ast)-\frac{(f_t(\x_{t+1})-f_t(\x_t^\ast))^2}{2\alpha\|\x_{t}-\x_t^\ast\|_2^2}\\
\leq&f_t(\x_{t})-f_t(\x_t^\ast)-\frac{(f_t(\x_{t+1})-f_t(\x_t^\ast))^2}{2\alpha D^2}.
\end{split}
\end{equation}
Moreover, we notice that
\begin{equation}
\label{OGD_eq5}
\begin{split}
&\sum_{t=1}^Tf_t(\x_t)-\sum_{t=1}^Tf_t(\x_t^\ast)\\
=&f_1(\x_1)-f_1(\x_1^\ast)+\sum_{t=1}^{T-1}\left(f_{t+1}(\x_{t+1})-f_{t+1}(\x_{t+1}^\ast)\right)\\
\overset{(\ref{lem1-eq1})}{\leq}&f_1(\x_1)-f_T(\x_T^\ast)+V_T+\sum_{t=1}^{T-1}\left(f_{t}(\x_{t+1})-f_{t}(\x_{t}^\ast)\right)\\
\overset{(\ref{OGD_eq4})}{\leq}&f_1(\x_1)-f_T(\x_T^\ast)+V_T+\sum_{t=1}^{T-1}\left(f_t(\x_{t})-f_t(\x_t^\ast)-\frac{(f_t(\x_{t+1})-f_t(\x_t^\ast))^2}{2\alpha D^2}\right).
\end{split}
\end{equation}
From the above inequality, we first have
\begin{equation}
\label{OGD_eq6}
\begin{split}
\sum_{t=1}^{T-1}(f_t(\x_{t+1})-f_t(\x_t^\ast))^2\leq 2\alpha D^2(f_1(\x_1)-f_T(\x_T)+V_T)\leq2\alpha D^2(M+V_T).
\end{split}
\end{equation}
Then, we further have
\begin{equation}
\label{OGD_eq7}
\begin{split}
\sum_{t=1}^Tf_t(\x_t)-\sum_{t=1}^Tf_t(\x_t^\ast)\overset{(\ref{OGD_eq5})}{\leq}&f_1(\x_1)-f_T(\x_T^\ast)+V_T+\sum_{t=1}^{T-1}\left(f_{t}(\x_{t+1})-f_{t}(\x_{t}^\ast)\right)\\
\leq&f_1(\x_1)-f_T(\x_T^\ast)+V_T+\sqrt{(T-1)\sum_{t=1}^{T-1}\left(f_{t}(\x_{t+1})-f_{t}(\x_{t}^\ast)\right)^2}\\
\overset{(\ref{OGD_eq6})}{\leq}&f_1(\x_1)-f_T(\x_T^\ast)+V_T+\sqrt{2\alpha D^2(T-1)(M+V_T)}\\
\leq&M+V_T+\sqrt{2\alpha D^2(T-1)(M+V_T)}
\end{split}
\end{equation}
where the second inequality is due to Cauchy–Schwarz inequality.

\end{proof}

\end{document}